\documentclass{article}
\usepackage{ijcai20}
\usepackage{paralist}
\usepackage{times}
\usepackage{soul}
\usepackage{url}
\usepackage[hidelinks]{hyperref}
\usepackage[utf8]{inputenc}
\usepackage[small]{caption}
\usepackage{graphicx}
\usepackage{amsmath}
\usepackage{amsthm}
\usepackage{booktabs}
\usepackage{algorithm}
\usepackage{algorithmic}
\usepackage{multirow}

\title{Modeling Global Semantics for Question Answering over Knowledge Bases}

\author{Peiyun Wu \and Yunjie Wu \and Linjuan Wu \and Xiaowang Zhang \and Zhiyong Feng\\
College of Intelligence and Computing, Tianjin University,  China}

\begin{document}
	
	\maketitle
	
	\begin{abstract}
		
Semantic parsing, as an important approach to question answering over knowledge bases (KBQA), transforms a question into the complete query graph for further generating the correct logical query. Existing semantic parsing approaches mainly focus on relations matching with paying less attention to the underlying internal structure of questions (e.g., the dependencies and relations between all entities in a question) to select the query graph. In this paper, we present a relational graph convolutional network (RGCN)-based model gRGCN for semantic parsing in KBQA. gRGCN extracts the global semantics of questions and their corresponding query graphs, including structure semantics via RGCN and relational semantics (label representation of relations between entities) via a hierarchical relation attention mechanism. Experiments evaluated on benchmarks show that our model outperforms off-the-shelf models.
	\end{abstract}
	
	\vspace*{-0.5cm}  
	\section{Introduction}
	Semantic parsing~\cite{berant13,YihCHG15} constructs a semantic parsing tree or equivalent query structure (called \emph{query graph}~\cite{Luo18}) that represents the semantics of questions. Semantic parsing based approaches effectively transform questions into logical queries where the reliability of logical querying can ensure the correctness of answering questions. The success of semantics parsing lies in representing the semantics of questions in a syntactic way so that it can better capture the intention of users~\cite{HuZZ18}.
	
Most existing semantic parsing approaches in question answering over knowledge bases (KBQA) focus on capturing the semantics of questions and query graphs. Some ``common'' relations occurring in both questions and query graphs are taken as core relations for measuring similarity together with some manual features~\cite{YihCHG15,bao16}. Another approach to detecting individual relation is presented in \cite{YuYHSXZ17} to improve the performance of matching questions with query graphs where each relation is represented by integrating its word-level and relation-level representations. \cite{Luo18} extends \cite{YuYHSXZ17} by improving the representation of questions with answering more complex questions. \cite{Sorokin18} enriches the semantics of all entities via all relations during the learning process by Gated Graph Neural Network~\cite{LiTBZ15} (GGNN). 
	
\begin{figure}[H]
		\centering
		\includegraphics[width=0.8\linewidth,height=3cm]{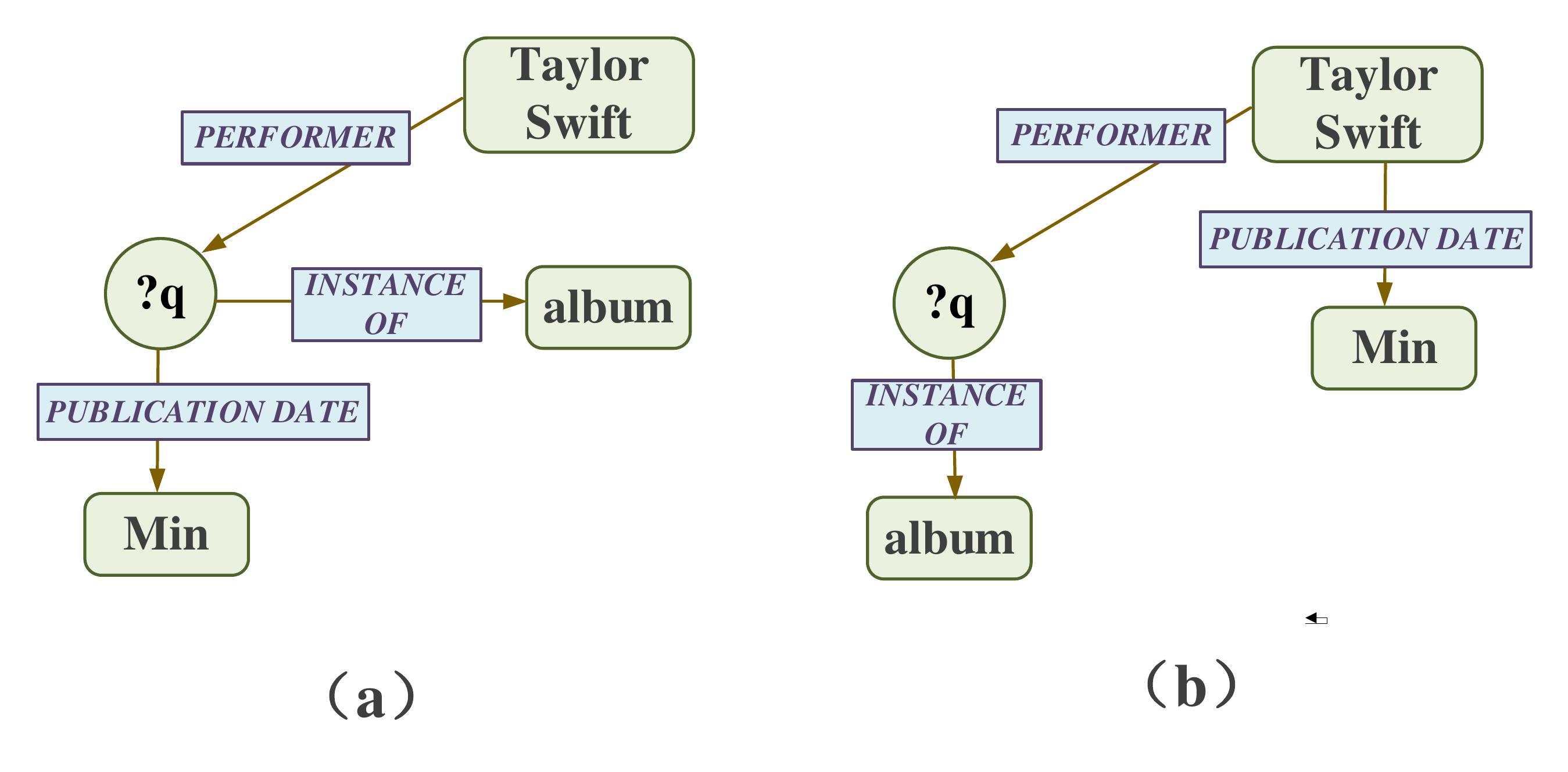}
		\vspace*{-0.4cm}
		\caption{(a) is the correct query graph which can find the correct answers in knowledge
			base, (b) is a wrong query graph with same relations but different structure in (a).}\label{fig:sg}
		\vspace*{-0.4cm}
	\end{figure}

However, the state-of-the-art semantic parsing approaches utilize relational semantics of query graphs with pay little attention to the structure semantics of a question. The structure semantics is an important part of the whole semantics of questions (e.g., Figure~\ref{fig:sg}), especially in complex questions where the complexity of a question often relies on its complicated structure. As a result, existing works only consider relational semantics cannot always perform complex questions better. So it is necessary to pay more attention to the structure semantics of questions together with relational semantics when semantic parsing in KBQA. However, to model multi-relational directed graphs with edge features remains an open problem. Therefore, it is not trivial to combine the two semantics of a question.

	%%%
	%We consider the combination of structure and relations semantics to be the global semantics of the question, which is not negligible for resolving KBQA and maximize performance of answering various questions. Nevertheless, query graph as a directed graph with labeled nodes and typed directed edges, it's difficult to simultaneous represent the relation semantics and the structure semantics.
	
	\vspace*{-0.4cm}
	 %% font size bigger
	
In this paper, based on a RGCN~\cite{Schlichtkrull18}, we propose a novel model gRGCN for global semantic parsing of KBQA.  In gRGCN,  we apply RGCN to extract structure semantics of query graphs by utilizing its capability of learning the better task-specific embeddings of multi-relational graphs. gRGCN focuses on extracting global semantics of both query graphs and relations. The main contributions are summarized as:
	\begin{compactitem}
		\item We propose a global semantic fusion method for structure semantics of query graphs extracted in RGCN integrated with relational semantics via enhanced fine-grained relations (\emph{word-level} and \emph{relation-level} representation) learning. 
		\item We introduce a hierarchical attention mechanism to represent word-level relations, which can discover latent implicit meaning and remove ambiguity in words.
		\item We introduce a syntactic-sequence-combined representation to encode questions and relation-attention-based RGCN layer to strengthen structure semantics. 
	\end{compactitem}
	
	\begin{figure*}[t]
		\centering
		\includegraphics[width=0.85\linewidth]{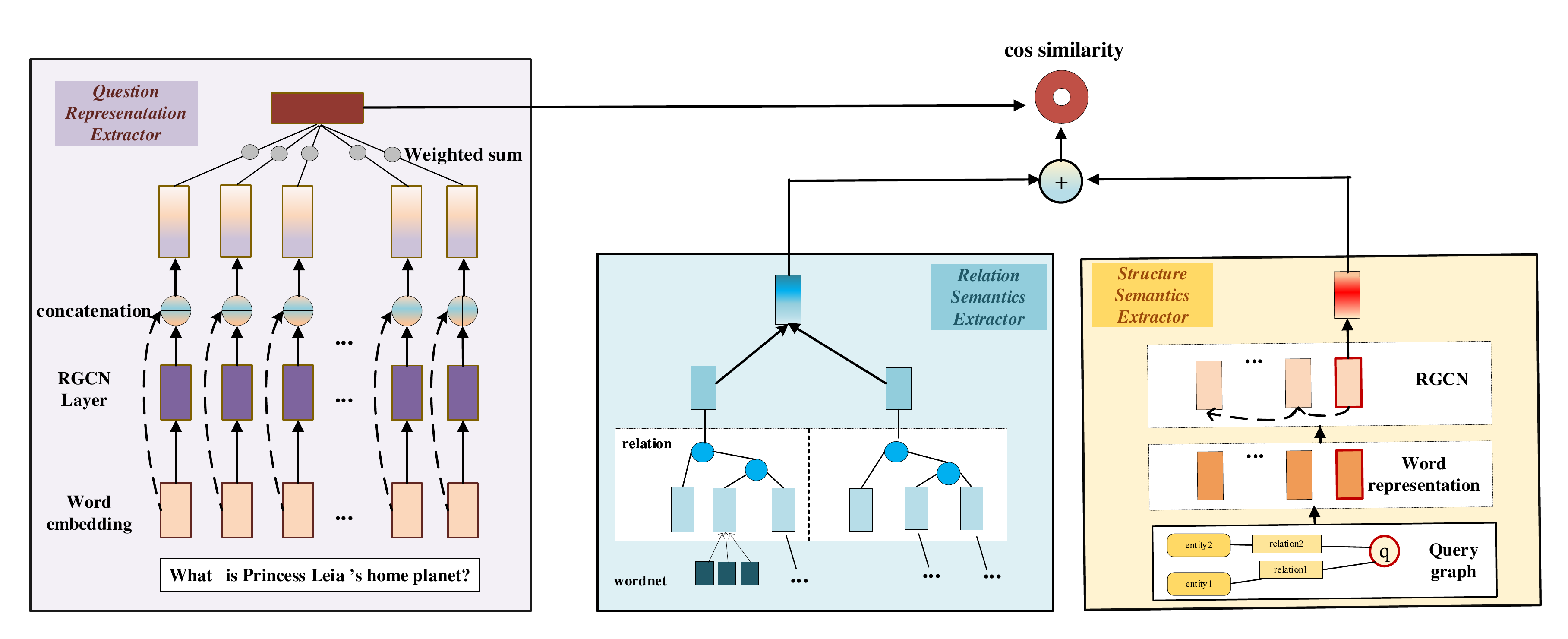}
		\caption{The Overview of gRGCN.} \label{fig2}
	\end{figure*}

\section{The Overview of Our Model}\label{sec:overview}
In this section, we introduce our gRGCN model for global semantic parsing in KBQA, and the overview framework is shown in Figure~\ref{fig2}.
	
Our gRGCN model consists of three modules, namely, \emph{question representation encoder}, \emph{structure semantics extractor}, and \emph{relational semantics extractor}.
\begin{itemize}
\item \textbf{Question Representation Encoder} transforms all questions input into their syntactic dependency tree as input of RGCN and then encodes the representations of all questions in RGCN with combining the sequence of words. The module is used to provide the condition of calculating the similarity of a question with its candidate query graphs together with the fused semantics of structure semantics and relational semantics. 
\item \textbf{Structure Semantics Extractor} embeds query graphs as a whole into a semantic space. The module is used to mainly extract the semantics of the whole structure of query graphs. In this module, we present a relation-attention-based RGCN layer for extracting global semantics to be fused with relational semantics extracted in the last module.
\item \textbf{Relational Semantics Extractor} embeds relations into a semantic space. The module is used to mainly extract the semantics of relations of query graphs left in the global semantics extractor. In this module, we present a hierarchical relational attention mechanism to reduce the noise of word-level relations via \emph{WordNet}~\cite{Miller95}.
	\end{itemize}

\section{Our Approach}\label{sec:approach}
In this section, we introduce our approach in the followings.
	
\subsection{Query Graph Generation} %%%%% section 3.1
We convert a question to query graphs further transformed into SPARQL queries for answering on KB. A query graph $G$ is a directed graph with labeled nodes and typed directed edges. All vertexes of $G$ are divided into three categories: question variable ($?q$), intermediate variables($v$), and KB entities. A $?q$-node represents the answer to a question. A $v$-node is either an unknown entity or an unknown value. An edge represents a relation between two vertexes.
	
In this paper, we improve the candidate graph generation method based on an heuristic algorithm~\cite{Schlichtkrull18} by further considering five kinds of semantic constraints: \textit{entity}, \textit{type}, \textit{temporal(explicit and inexplicit time)}, \textit{order}, and \textit{compare}.
	
\subsection{Question Representation Encoding} %%%% section 3.2
This module encodes the syntactic structure of a question via RGCN with combining with the sequence of all words occurring in the question to a final vector as the representations of the question.
	
Given a question $q =(w_1, \ldots, w_n)$, the syntactic dependency tree (a graph structure) of $q$, denoted as $G_q$, is of form $(V_q, E_q)$ where $V_q = \{w_1,\ldots, w_n\}$ (a vertex set) is a set of words and $E_q \subseteq V_q \times V_q$ (a edge set) is a set of dependency-relation over $\{w_1,\ldots,w_n\}$. Edges are classified into two classes: \emph{self-loop} and \emph{Head to dependent}.
	
Let $W_{\mathrm{glove}} \in \mathbf{R}^{\left| V \right|\times d}$ be the word embedding matrix where each row vector represents the embedding of a word, $d$ is the dimension of embedding. For each vertex $w_i \in V_q$, $\overrightarrow{h}_{w_i}$ is assigned to the corresponding vertex $w_i$ as its embedding in $V_q$ ($i=1,2,\ldots,n$). 
	
Firstly, we initialize the embedding of $V_q$ via its all vertexes' embedding and compute the average of word embeddings as $\overrightarrow{E}_{\mathrm{avg}}$.
	
Secondly, given the intial embedding of $V_q = \{\overrightarrow{h}^{(0)}_{w_1},...,\overrightarrow{h}^{(0)}_{w_n}\}$, we apply RGCN encoder to generate its hidden representation. In RGCN, the forward-pass update of each vertex $w_i$ in $V_q$ is formalized in the following:
	\begin{equation*}
	\overrightarrow{h}_{w_i}^{(l+1)} = \mathrm{ReLU} \left ( \sum_{r \in E_q}\sum_{w \in N^r_{w_i}} \frac{1}{\left|N^r_{w_i} \right|} W_r \overrightarrow{h}_{w}^{(l)}+W_0 \overrightarrow{h}_{w_i}^{(l)}\right).
	\end{equation*}
	Here $l$ is a layer, $N^r_{w_i}$ is the set of all $r$-neighbors of $w_i$ in $G_q$. Note that $W_0$ and $W_v$ are weighted matrixes to be learned. 
	
To access both sequence and syntactic information, we concatenate the word embedding of $q$ and the output of RGCN layers to calculate the final result. The attention weight of the $j$-th token (i.e., ${a_j}$) is calculated based on the following formula: let $M$ is a weight matrix,
	\begin{equation*}
	{a_i}=\mathrm{softmax}\left (\overrightarrow{E}_{\mathrm{avg}} \cdot M\cdot [\overrightarrow{h}^{(0)}_{w_i} \oplus \overrightarrow{h}^{(l+1)}_{w_i}]\right),~\forall~ i\in n.
	\end{equation*}

	Finally, we present a fully-connected layer and the ReLU non-linearity to obtain the final representation $h_q$ of question $q$ in the following equation:
	\begin{equation*}
	\overrightarrow{h}_{q} = \mathrm{ReLU} \left ( W \cdot (\Sigma^n_{i=1}~{a_i}[\overrightarrow{h}^{(0)}_{w_i}\oplus\overrightarrow{h}^{(l+1)}_{w_i}]) + \overrightarrow{b}\right).
	\end{equation*}
	
\subsection{Structure Semantics Extracting}%%%%% section 3.3
This module computes the representation of the structure of a query graph. We first formalize query graphs.
	
Let $V_e$ be a set of entites and $V_r$ be a set of relations. A query graph $G$ on $V_e$ and $V_r$ is a quad of $(N_G, R_G, \lambda, \delta)$ where $N_G$ is the set of vertexes, $R_G$ is the set of edges, $\lambda: N_G \to V_e$ assigns each vertex to an entity, and $\delta: R_G \to V_r$ assigns each edge to a relation. 
	
Firstly, we initialize hidden states $\overrightarrow{h}^{(0)}_{v_i}$ for node $v_i$ by calculating them in relation-attention-based RGCN layer in the following way:
	\begin{equation*}
	\overrightarrow{h}_{v_i}^{(l+1)} = \mathrm{ReLU} \left ( \sum_{r \in R_G}\sum_{v \in N^r_{v_i}} ({a_r} W_v \cdot \overrightarrow{h}_{v}^{(l)}+W_0 \cdot \overrightarrow{h}_{v_i}^{(l)})\right).
	\end{equation*}
	Here $l$ is a layer, $N^r_{v_i}$ is the set of all $r$-neighbors of $v_i$ in $G$. Note that $W_0$ and $W_v$ are weighted matrixes to be learned. $a_r$ is the attention weight of the edge $r$ to give high weight to relations more relevant to the question in query graphs :
	\begin{equation*}
	{a_r}= \mathrm{softmax} \left (\overrightarrow{E}_{\mathrm{avg}} \cdot \overrightarrow{{r}_i}^{T}\right),~\forall~ {r}_i\in R_G.
	\end{equation*}
	
	Finally, we obtain $\overrightarrow{h}_{\mathrm{structure}}$ as the structure semantic representation of $G$ by taking the representation of answer node.
	\begin{figure}
		\centering
		\includegraphics[width=0.8\linewidth]{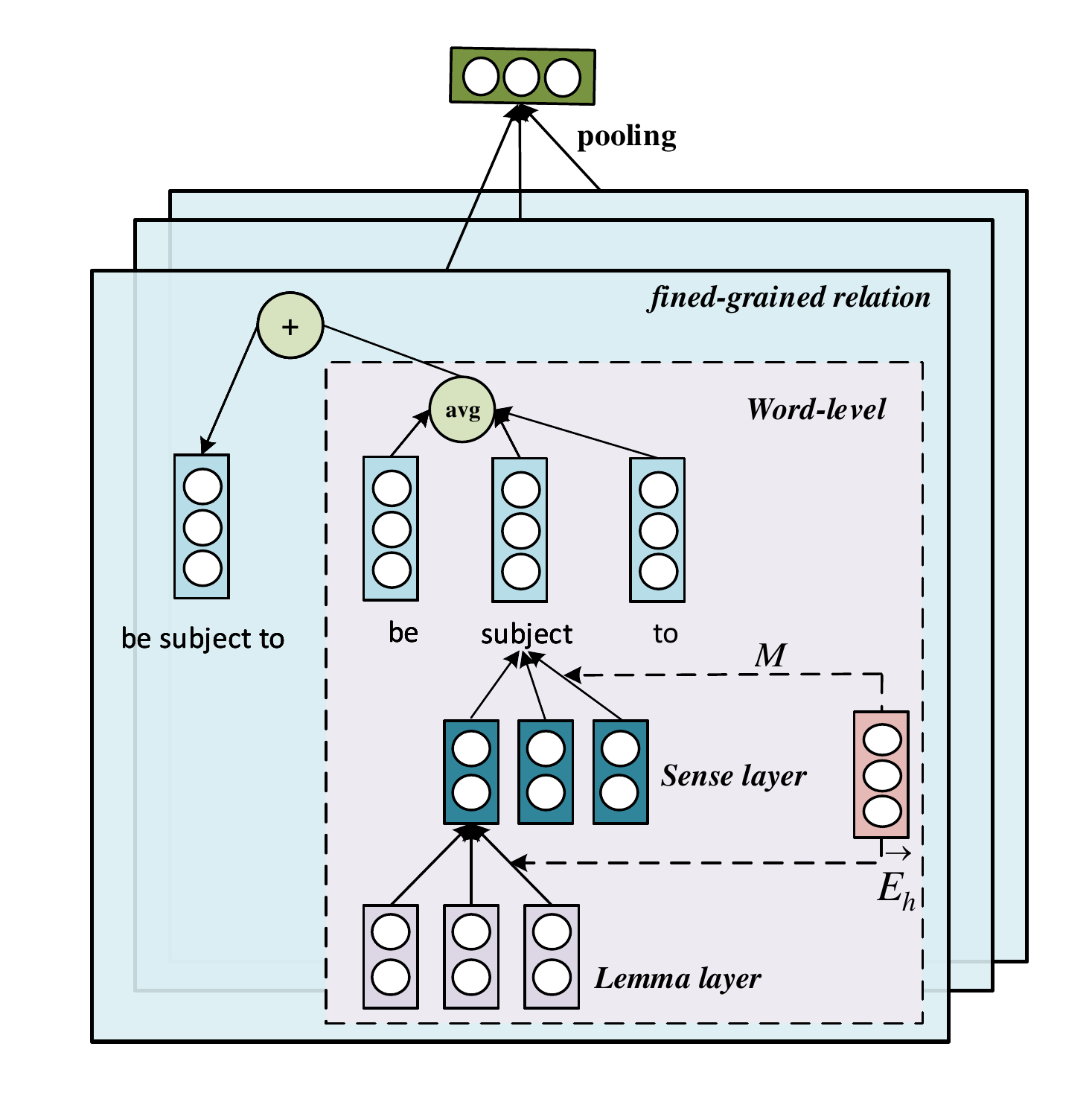}
		\caption{Local Semantics Extractor} \label{fig3}
	\end{figure}
	\vspace*{-0.4cm}    
	\subsection{Relational Semantics Extracting}%%% section 3.4
	To utilize relational semantics both questions and query graphs, we compute the representation of all relations in \emph{word-level} and \emph{relation-level}.
	
	Firstly, we define a \emph{relational embedding matrix} as follows: let $d$ be a dimension, $W_{\mathrm{rel}}\in \mathbf{R}^{\left| V_r \right|\times d}$.
	%\begin{equation}\label{equrel}
	%W_{\mathrm{rel}}\in \mathbf{R}^{\left| V_r \right|\times d}.
	%\end{equation}
	
	Given a query graph $G = (N_G, R_G, \lambda, \delta)$ on $V_e$ and $V_r$, we define \emph{relation-level} relations (e.g. \emph{``posistion held''}) in $G$, denoted by $\mathcal{R}^{\mathrm{whole}}$, as a sequence of vectors as follows: let $\lambda(R_G) = \{r_1, \ldots, r_n\}$, $\mathcal{R}^{\mathrm{whole}}_{1:n} := \{\overrightarrow{r}_1, \ldots, \overrightarrow{r}_n\}$.
	%\[
	%\mathcal{R}^{\mathrm{whole}}_{1:n} := \{\overrightarrow{r}_1, \ldots, \overrightarrow{r}_n\}.
	%\]

	We define word-level relation $\mathcal{R}^{\mathrm{word}}$, as a sequence of vectors by applying $W_{glove}\in \mathbf{R}^{\left| V\right|\times d}$ to map each word $w_{ij}$ in $r_i$ to its word embedding $\overrightarrow{w}_{ij}$ as follows: $\mathcal{R}^{\mathrm{word}}_{r_i} := \{\overrightarrow{w}_{i_1}, \ldots, \overrightarrow{w}_{i_m}\}$, where $\{w_{i_1}, \ldots, w_{i_m}\}$ is a set of all words occurring in $r_i$ (e.g.{\emph{``posistion''}, \emph{``held''}}).
	%\[
	%\mathcal{R}^{\mathrm{word}}_{r_i} := \{\overrightarrow{w}_{i_1}, \ldots, \overrightarrow{w}_{i_m}\},
	%\]

\begin{figure}[H]
		\centering
		\includegraphics[width=0.7\linewidth]{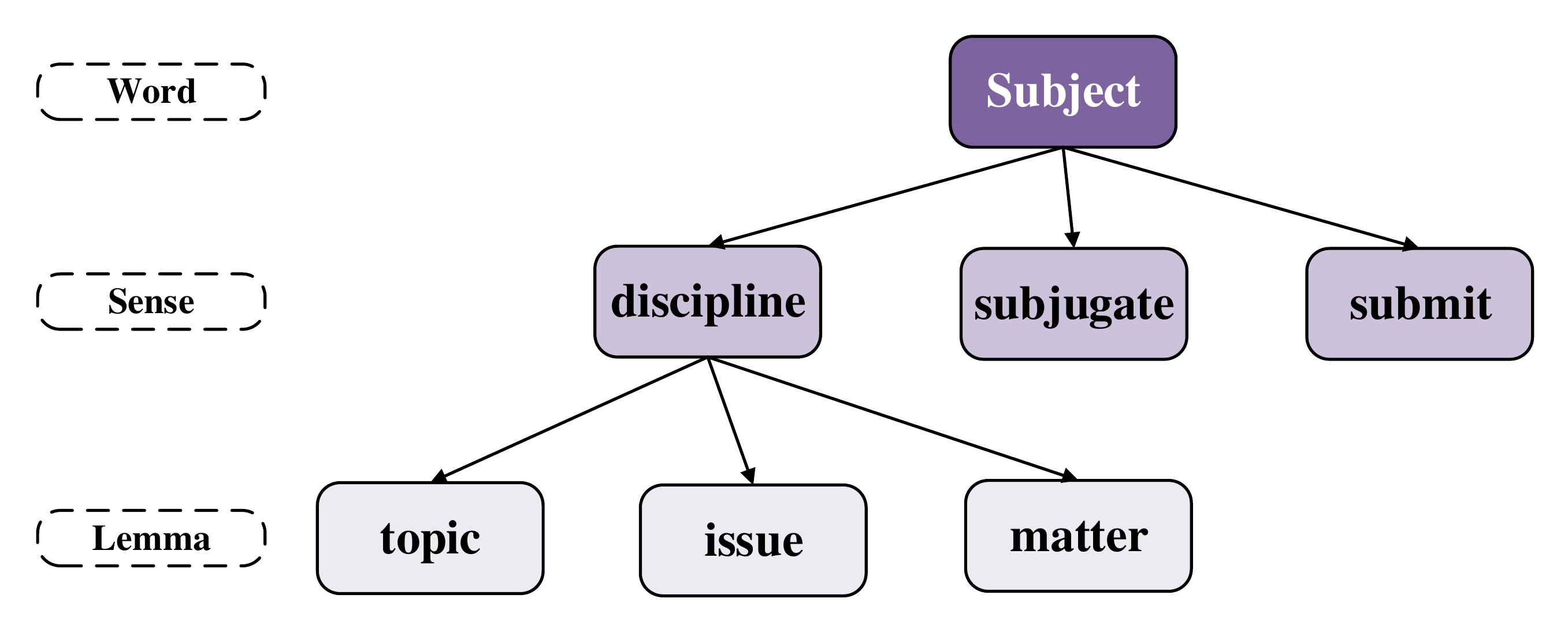}
		\vspace*{-0.1cm}
		\caption{ The structure of ``subject'' in \emph{Wordnet}}\label{fig4}
	\end{figure}

	There is some unavoidable noise of fine-grained relations. For instance, for relation type "\emph{be subject to}," its word-level word may lose the global relation information, as "\emph{subject}" may have different meanings shown in Figure~\ref{fig4}.

	To reduce noise, we introduce an external linguistic knowledge \emph{WordNet}~\cite{Miller95}, where a word possibly has multiple senses, and a sense consists of several lemmas.
	\vspace*{-0.2cm}
	For each word $w_{ij}$ in a relation $r_i$, we use $S{w_{ij}}=\{\overrightarrow{w}_{ij1}, \ldots, \overrightarrow{w}_{ijk}\}$ to denote all sense vectors in $w_{ij}$ and $L{w_{ijk}}=\{\overrightarrow{w}_{ijk1}, \ldots, \overrightarrow{w}_{ijkz}\}$ to denote all lemma vectors in sense $w_{ijk}$. The vector of sense $\overrightarrow{w}_{ijk}$ is defined as follows:
	\vspace*{-0.2cm}
	\begin{equation*}
	\overrightarrow{w}_{ijk}= \sum^z_{y=1} {a_{ijky}} \overrightarrow{w}_{ijky}, ~\text{where}
	\end{equation*}
	\vspace*{-0.2cm}
	\begin{equation*}
	{a_{ijky}}=\mathrm{softmax} \left (\tanh \left (\overrightarrow{\mathrm{avg}_q}\cdot \overrightarrow{w}^{T}_{ijky}\right )\right),~~\forall~y\in {(1,\dots,z)}
	\end{equation*}
	
	The vector of word $w_{ij}$ is defined as follows:
	\begin{equation*}
	\overrightarrow{w_{ij}}= \sum^k_{c=1} {a_{ijc}} \overrightarrow{w_{ijc}}, ~\text{where},~~\forall~c\in {(1,\dots,k)}
	\end{equation*}
	\vspace*{-0.3cm}
	\begin{equation*}
	{a_{ijc}}= \mathrm{softmax} \left(W \cdot \left(\tanh \left (\overrightarrow{\mathrm{avg}_q}\cdot \overrightarrow{w}^{T}_{ijc}\right )\right)+\overrightarrow{b}\right)
	\end{equation*}
	
	Then, we obtain the \emph{fine-grained relations embedding of} $G$, denoted by $\mathcal{R}^{\mathrm{fine}}$. Based on fine-grained word embedding of a relation $r_i$, we define a vector as follows:
	\begin{equation*}
	\mathcal{{R}^{\mathrm{fine}}_G}[i]= \frac{1} m \sum^m_{j} \mathcal{R}^{\mathrm{WORD}}_{r_{i}}[j] + \mathcal{R}^{\mathrm{WHOLE}}[i],~~\forall~i\in {(1,\dots,n)}.
	\end{equation*}
	
	We employ max pooling over the $\mathcal{R}^{\mathrm{fine}}$ to get the final relational semantics $\overrightarrow{h}_{\mathrm{relational}}$.
	
	Finally, we present a linear combination to fuse structure and relational semantics via ReLU as the global semantics: 
	%$\overrightarrow{h}_{\mathrm{whole}}$.
	\begin{equation*}
	\overrightarrow{h}_{\mathrm{whole}} = \mathrm{ReLU}(W^{{\tiny \mathrm{T}}}(\overrightarrow{h}_{\mathrm{relational}}+ \overrightarrow{h}_{\mathrm{structure}})+\overrightarrow{b}).
	\end{equation*}

	\section{Experiments and Evaluation}%%%% section 5
	In this section, we evaluate our approach on the following experiments: an overall contrast experiment with baselines and ablation study experiments of our three different modules.
	
	\renewcommand{\arraystretch}{1.2}
	\begin{table*}[h]  
		\centering
		\caption{Overall Average Results over Wikidata}\label{tab:All}  
		\begin{tabular}{l|l|l|l|lll|l|l|l|}
			\hline
			\multicolumn{1}{c|}{\multirow{2}{*}{\textbf{Model}}} & \multicolumn{3}{c|}{\textbf{WebQSPS}}                          & \multicolumn{3}{c|}{\textbf{QALD-7}}                                  & \multicolumn{3}{c|}{\textbf{CompQ}}                         \\ \cline{2-10} 
			\multicolumn{1}{c|}{}                & \multicolumn{1}{c|}{Precision} & \multicolumn{1}{c|}{Recall} & \multicolumn{1}{c|}{F1} & \multicolumn{1}{c|}{Precision}    & \multicolumn{1}{c|}{Recall}     & \multicolumn{1}{c|}{F1} & \multicolumn{1}{c|}{Precision} & \multicolumn{1}{c|}{Recall} & \multicolumn{1}{c|}{F1} \\ \hline
			STAGG(2015)                     & 0.1911            & 0.2267           & 0.1828         & \multicolumn{1}{l|}{0.1934}     & \multicolumn{1}{l|}{0.2463}     & 0.1861         & 0.1155                & 0.1481              & 0.1108            \\ \cline{2-10} 
			
			Yu et al.(2017)                   &0.2144                & 0.2548              & 0.2006            & \multicolumn{1}{l|}{0.1972}     & \multicolumn{1}{l|}{0.2103}     & 0.1923         &  0.1297               & 0.1675              & 0.1291            \\ \cline{2-10} 
			Luo et al.(2018)                   & 0.2374             & 0.2587           & 0.2252         & \multicolumn{1}{l|}{0.2117}     & \multicolumn{1}{l|}{0.2438}     & 0.2016         & 0.1331               & 0.2118         & 0.1317            \\ \cline{2-10} 
			GGNN(2018)                      & 0.2686             & 0.3179           & 0.2588         & \multicolumn{1}{l|}{0.2176}     & \multicolumn{1}{l|}{0.2751}     & 0.2131         & 0.1297               & 0.1481             & 0.1285            \\ \cline{2-10} 
			\hline
			gGCN (our)                       & 0.2713             & 0.3291           & 0.2631         & \multicolumn{1}{l|}{0.2334}     & \multicolumn{1}{l|}{0.3109}     & 0.2437         & 0.1267                &  0.2244             & 0.1441            \\ \cline{2-10} 
			gRGCN (our)                        & \textbf{0.3009}        & \textbf{0.3475}       & \textbf{0.2910}     & \multicolumn{1}{l|}{\textbf{0.3133}} & \multicolumn{1}{l|}{\textbf{0.3541}} & \textbf{0.3024}     & \textbf{0.1400}           & \textbf{0.2558 }          & \textbf{0.1519}        \\ \hline
		\end{tabular}
	\end{table*}
	\vspace*{-0.2cm}  
	\subsection{Experiments Setup}
	\paragraph{Knowledge Bases} We select two representive KBs:
	\begin{compactitem}
		\item \emph{Wikidata}: A collaborative KB developed by ~\cite{wikidata} contains more than 40 million entities and 350 million relation instances. We use the full Wikidata dump\footnote{\url{https://www.wikidata.org/wiki/Special:Statistics}} and host it with Virtuoso engine\footnote{\url{http://virtuoso.openlinksw.com/}}.
		\item \emph{FB2M}: The KB collected by~\cite{simple}, as a subset of Freebase, consists of 2 million entities and 10 million triple facts. FB2M is a famous KB for creating many QA datasets particularly simple question datasets.
	\end{compactitem}
	
	\paragraph{\textbf{QA Datasets}} We select four popular datasets as follows:
	\begin{compactitem}
		\item {\em ComplexQuestion} (CompQ): The dataset consisting of 2,100 complex questions is developed by ~\cite{bao16} to provide questions with structure and expression diversity. To support answering over Wikidata~\cite{wikidata}, we slightly revise it by mapping answers to Wikidata answer IDs instead of Freebase IDs since Freebase was discontinued and is no longer up-to-date, including unavailability of APIs and new dumps~\cite{Sorokin18}. %% Specifically, we generate positive semantic graphs for 1200 questions out of 2100 from CompQ and then take 400 of it as test set in order to get answers to all questions from Wikidata always by overtaking the gap between Freebase and Wikidata.
		
		\item {\em WebQSP-WD} (WebQSPS): The dataset collected by ~\cite{Sorokin18} contains 712 real complex questions (in total 3913). WebQSPS is a corrected version of the common benchmark WebQSP\cite{berant13} data set for supporting Wikidata.
		
		\item {\em QALD-7}: The small dataset consisting of 42 complex questions and 58 simple questions collected by~\cite{qald} to be mainly used to test KBQA models. QALD-7 is specially designed for the KBQA task over Wikidata.
		
		\item {\em SimpleQuestion} (SimpQ): The dataset developed by ~\cite{simple}, as a popular benchmark of KBQA, contains over 100K questions w.r.t. FB2M. To evaluate the generalization ability of our model, we additionally use SimpQ in our complementary experiment.  
	\end{compactitem}

	\paragraph{Implementation Details}
	In training, we adopt \emph{hinge loss} which are applied in~\cite{Luo18,Sorokin18}. Formally, let $q$ be a question and $C$ be the query graph set of $q$ where $C$ contains all positive graph $g^+$ and negative graph $g^-$ of $q$. 
	\[
	\mathcal{L} = \max\sum_{g\in C}(0,(\lambda-\cos(\vec{q}, \vec{g}_+)+\cos(\vec{q}, \vec{g}_-))). %%% to be checked 1/18
	\]
	
	In our experiments, We use S-MART~\cite{YangC15} as our entity linking tool, GloVe~\cite{PenningtonSM14} word vectors with dimensions of 100 is employed to initializing word embeddings and Adam optimizer~\cite{KingmaB14} is applied to train the model with a batch size of 64 at each epoch. Moreover, 
	we design a multiple-layer RGCN with a fully-connected layer with dropout=0.2 to calculate the representation of questions and the structure of query graphs (three-layer for questions and two layers for graphs). Finally, \emph{SpaCy}\footnote{https://spacy.io/.} is used to parse syntactic dependency, and Deep Graph Library (DGL\footnote{https://www.dgl.ai/pages/about.html.}) is used to transfer query graphs into DGL graph objects.

	\subsection{Baselines}
	We introduce four popular models and one variation of gRGCN as baselines as follows:
	\begin{compactitem}
		\item STAGG (2015)~\cite{YihCHG15}: The model scores query graphs by some manual features and the representation of the core relation. It's a re-implemented by~\cite{Sorokin18}.
		\item Yu et al. (2017)\cite{YuYHSXZ17}: The model encodes questions that are encoded by residual BiLSTM and compute similarity with the pooling of fined-grained relations. We re-implement it by adding a graph generating process in order to support complex questions.
		\item Luo et al. (2018)\cite{Luo18}: The model represents the corresponding multi-relational semantics of complex questions.
		\item GGNN (2018)~\cite{Sorokin18}: The model encodes questions by Deep Convolutional Neural Network (DCNN), and query graphs are encoded based on Gated Graph Neural Network (GGNN).
		\item gGCN (our): The model is obtained from gRGCN by replacing RGCN with Graph Convolutional Network(GCN)~\cite{Kipf17} where GCN ignores the relational type .
	\end{compactitem}
	
	The four models as our baselines are representative in various mechanisms: STAGG (2015) is based manual features in characterizing the structure of query graphs; Yu et al. (2017) is based on the max-pooling of fined-grained relational semantics; Luo et al. (2018) is based on the sum fined-grained relational semantics; GGNN (2018) enriches the semantics of each entity by the average of relations semantics. Besides STAGG(2015), there are many works mainly on constructing query graphs over Freebase, such as \cite{bao16,HuZZ18}. Since their codes are not accessible, we will discuss them in the related works. %%% to be checked 1/18
	\vspace*{-0.2cm}  
	\subsection{Overall Results of Our Approach}
	For evaluating our approach on complex questions, we select Wikidata as KB, CompQ, WebQSPS, and QALD-7 as datasets and precision, recall, and F1-score as metrics.
	
	The experimental results are shown in Table~\ref{tab:All}, where we take the average results of all questions in a dataset as the final result. Note that our results are different from those original results released in baselines over Freebase.
	
	By Table~\ref{tab:All}, we show that gRGCN outperforms all datasets and all metrics.
	\begin{compactitem}
		\item gRGCN achieves $59.2\%$, $12.4\%$ higher F1-score compared to STAGG and GGNN on WebQSPS; $62.4\%$, $41.9\%$ higher F1-score compared to STAGG and GGNN in QALD-7 dataset; $37.1\%$, $18.2\%$ higher F1-score compared to STAGG and GGNN on CompQ. So we can conclude that the global semantics performs better than all baselines without considering the global semantics.    
		\item The five models: Yu et al. (2017), Luo et al. (2018), GGNN (2018), gGCN (our), and gRGCN (our) achieves $59.2\%$, $12.4\%$ higher F1-score compared to STAGG (2015) on the three datasets. In short, we can show that end-to-end neural network frameworks perform better than models with manual features.  %%% checking value
		\item gRGCN achieves $10.2\%$, $24.1\%$, $5.4\%$ higher F1-score compared to gGCN on the three datasets. Hence, we can conclude that relational types as a part of a relational structure can also improve the performance of KBQA.
	\end{compactitem}
	
	Therefore, the experiments show that our approach can extract the global semantics of complex questions, which is helpful in improving the performance of KBQA.
	
	\subsection{Robustness of Our Approach}
	
	Simple questions still contain a little structural information, such as the linear order of the subject entity and the object entity. We complementary evaluate our approach to simple questions to analyze its robustness. This experiment selects FB2M as KB, SimpQ as dataset, and six popular baselines.
	\begin{table}[h]
		\centering
		\vspace*{-0.3cm}  
		\caption{Results on Simple Question}\label{tab:Simple} 
		\vspace*{-0.2cm}    
		\begin{tabular}{|l|l|}
			\hline
			\multicolumn{1}{|c|}{\textbf{SimpQ}} & \multicolumn{1}{c|}{\textbf{Accuracy}} \\ \hline
			Yin et al.(2016)               & 0.683              \\ \hline
			Bao et al.(2016)               & 0.728              \\ \hline
			Lukovnikov et al.(2017)            & 0.712              \\ \hline
			Luo et al.(2018)               & 0.721                 \\ \hline
			Mohammed et al.(2018)             & 0.732              \\ \hline
			Huang et al.(2019)              & \textbf{0.754}              \\ \hline
			\textbf{gRGCN (our)}                     & 0.739               \\ \hline
		\end{tabular}
		\vspace*{-0.2cm}  
	\end{table}
	By the experimental results shown in Table~\ref{tab:Simple}, our approach still keeps a competitive score and the global semantics improves the performance of simple questions.    
	
	Note that Huang et al. (2019), with a lot of simple question optimization, achieves $2\%$ slightly better our gRGCN . Indeed, Huang et al. (2019) recovers the question's head entity, predicate, and tail entity representations in the KG embedding spaces. In this sense, Huang et al. (2019) consider a structure-like semantics to improve the performance of simple question. In other words, Huang et al. (2019) also verifies the effectiveness of our idea.
	\vspace*{-0.2cm}  
	\subsection{Ablation Study}
	In this subsection, we will pay attention to analyze the effectiveness of the proposed question representation and the proposed query graph representation in our model. For the ablation study, we use the F1 score as our metrics and perform experiments on CompQ, WebQSPS, and QALD-7 w.r.t. complex questions.
	
	\paragraph{\textbf{Question Representation}}  
	To analyze the effectiveness of our proposed question representation encoder, this experiment compares gRGCN to gRGCN with substituting the question encoder in each baseline for our encoder. 
	
	The results are shown in Table~\ref{tab:QR}, where the column of QR of Baselines is gRGCN equipping with the corresponding question encoder and gRGCN$^{-}$ denote a variant of gRGCN without concatenating question sequences. We use gRGCN$^{-}$ to analyze the effectiveness of our proposed question encoder further, only considering those information of questions also considered in baselines fairly. 
	\begin{table}[H]
		\centering
		\vspace*{-0.2cm}  
		\caption{Ablation Results on Question Representation}\label{tab:QR}
		
		\vspace*{-0.2cm}  
		\begin{tabular}{|l|l|l|l|}
			\hline
			QR of Baselines & WebQSP & QALD & CompQ\\
			\hline
			Yu et al.(2017)  & 0.2784  & 0.2464 & 0.1400 \\
			\hline
			DCNN(STAGG,GGNN)    &0.2810 & 0.2511 & 0.1432 \\
			\hline
			Luo et al.(2018)   & 0.2844 & 0.2683 & 0.1483 \\
			\hline
			\hline
			gRGCN$^{-}$ (our)  & \textbf{0.2880}  & \textbf{0.2869} & \textbf{0.1499} \\
			\hline
			gRGCN (our)  & \textbf{0.2910}  & \textbf{0.3024} & \textbf{0.1519} \\
			\hline
		\end{tabular}
		\vspace*{-0.2cm}  
	\end{table}
	
	By Table~\ref{tab:QR}, we can show that our proposed question representation outperforms question representations of all baselines over all datasets. 
	\begin{compactitem}
		\item Firstly, our question encoder performs best compared to the other models. Specifically, gRGCN betters Luo et al. (2018), with a $2.3\%$, $12.7\%$, and $2.4\%$ improvement on three datasets.
		\item Secondly, our question encoder performs best compared to the other models without concatenating question sequences. Specifically, gRGCN$^{-}$ betters Luo et al. (2018) with a $1.3\%$, $6.9\%$, and $1.1\%$ improvement on three datasets.
		\item Thirdly, the question sequence is useful to improve the performance question presentation. Specifically, gRGCN betters gRGCN$^{-}$ with a $1.0\%$, $5.4\%$, and $1.0\%$ improvement on three datasets.
	\end{compactitem}
	
	\paragraph{\textbf{Graph Representation}} 
	To analyze the effectiveness of our proposed graph representation encoder, we consider two types of relations: fine-grained relation (f-g) and wordnet, two types of structures: attention and non-attention (non-att). Note that gRGCN considers all relations and structures.
	
	The experimental results on the three datasets are shown in Table~\ref{tab:graph}, where there are 18 cases, including 6 cases without structure.
	%%% to be added
	%In the structure of Table~\ref{tab:graph}, we use "-" to denote the graph structure is not encoded and we only consider relation, use "none" to denote the graph structure is considered without applying attention mechanism, and "attention" to denote that both structure and attention mechanism are considered. In relation of Table~\ref{tab:graph}, "f-g" represents simple fined-grained relations representation, "f-g/wordnet" represents our hierarchical relation attention mechanism.
	
	\begin{table}[H]
		\centering
		\vspace*{-0.2cm}
		\caption{Ablation Results on Graph Representation}\label{tab:graph}
		\begin{tabular}{|l|l|l|l|l|l|l|}
			\hline
			Relation& Structure &WebQSP & QALD & CompQ\\     
			\hline
			f-g & -   &0.2101 & 0.2081 & 0.1320 \\
			\hline
			f-g & non-att  & 0.2614 & 0.2472 & 0.1456 \\     
			\hline
			f-g & attention  & 0.2810 & 0.2690 & 0.1471 \\
			\hline
			f-g/wordnet & - & 0.2304  & 0.2501& 0.1368 \\
			\hline
			f-g/wordnet & non-att & 0.2750  & 0.2711& 0.1483 \\
			\hline
			f-g/wordnet & attention & 0.2910  & 0.3024 & 0.1519 \\
			\hline
		\end{tabular}
		\vspace*{-0.2cm}
	\end{table}
	
	By Table~\ref{tab:graph}, we can show that our proposed graph representation outperforms graph representations of all baselines over all datasets. 
	\begin{compactitem}
		\item Firstly, the structure of query graphs is verified to improve the performance of graph presentation without considering structure. Specifically, non-attention achieve $24.4\%$, $18.8\%$, $10.3\%$ improvement on three datasets, and even when further considering wordnet-relation, it still performs with a $19.4\%$, $8.3\%$, $8.4\%$ improvement. %% 
		\item Secondly, the wordnet-relation of query graphs is verified to improve the performance graph presentation without considering the wordnet-relation. Specifically, wordnet-relation performs with a $3.5\%$, $12.4\%$, $10.7\%$ improvement on three datasets, and even when further considering attention-structure, it still performs with a $3.5\%$, $12.3\%$, $3.3\%$ improvement. %% 
		\item Thirdly, the attention-structure of query graphs is verified to improve the performance graph presentation without considering the attention-structure. Specifically, attention-structure performs with a $24.4\%$, $18.8\%$, $10.3\%$ improvement on three datasets, and even when further considering wordnet-relation, it still performs with a $19.4\%$, $8.3\%$, $8.4\%$ improvement. %% 
	\end{compactitem}
	
	Therefore, we can conclude that the structure presented in our graph representation as a global semantics is useful to improve the performance question presentation as well as wordnet-relation.
	\vspace*{-0.2cm}  
	\subsection{Error Analysis}
	We randomly analyze the major causes of errors on 100 questions, which returned incorrect answers.  
	
	\textbf{Semantic ambiguity and complexity (42\%)}: Due to the limitation of semantic graph generation, we can't generate graph space for some question since their complexity and semantic ambiguity. 
	
	\textbf{Correct graphs scored low (16\%)}: This type of error occurred when the model failed to extract the global semantics of questions and graphs. Although the correct graphs existed in the candidate set, we failed to select it.
	
	\textbf{Entity linking error (9\%)}: This error occurred due to the failure to extract the appropriate entities from the given question. And subsequently, we generate incorrect semantic graphs. We could correct questions by replacing the wrong entities.
	
	\textbf{Dateset and KB error (33\%)}: This error occurred since the incompatibility between wikidata and Freebase ID in datasets or the defects of datasets. The test datasets contain many open question whose answers may be partially or completely unstored in Wikidata.
	\vspace*{-0.2cm}  
	\section{Related Works}
	The most popular approaches proposed for the KBQA task can be roughly classified into two categories: \emph{semantic parsing} and \emph{information retrieval}.

	\paragraph{\textbf{Semantic Parsing}} 
	Semantic parsing (SP) based approaches use a formal representation to encode the semantic information of questions for obtaining answers on KB~\cite{berant13,ReddyTCKDSL16,ZhangCXW19}. The traditional SP-based methods~\cite{KwiatkowskiZGS11,KrishnamurthyM12,CaiY13} mainly rely on schema matching or hand-crafted rules and features.
	
	Recently, neural networks (NN) based approaches have shown great performance on KBQA. Different from \cite{cui17} and \cite{Abujabal17} tried to extract questions intention by a large number of templates automatically learned from KB and QA corpora, SP+NN employed neural network to encode the semantics of questions and query graphs, and then select the correct query graphs of given questions in the candidate set for querying later~\cite{bao16,Luo18,HuZZ18}. \cite{Luo18} encoded multiple relations into a uniform query graph to capture the fined-grained relations. ~\cite{Sorokin18} used GGNN to enrich the semantics of all entities with tackling the limited complex questions. Differently, our approach integrates both structure and enhanced relational semantics of query graphs. 
	
	~\cite{HuZZ18} ~\cite{DingHXQ19} pay more attention to query graph generation and use an SVM rank classifier and a combinational function to rank candidate graphs respectively, which is not an encode-and-compare
	framework as our model. ~\cite{SunHC18} uses an end-to-end query graph generation process via an RNN model, and it aims to search the best action sequence. However, we are focus on learning global semantics of the query graph.
	
	\paragraph{\textbf{Information Retrieval}}
	Information retrieval (IR) based approaches retrieve the candidate entities (as answers) from KB using semantic relation extraction~\cite{Xu_IR2016,DongWZX15,WuHWZZYC19,YihRMCS16,SunDZMSC18,LanW019}.Those approaches map answers and questions into the same embedding space, where we could retrieve the answers independent of any grammar, lexicon, or other semantic structure. Our approach belongs to the SP-based method and encoded semantic information of question into a query graph for querying on KB.
	
	\vspace*{-0.2cm}  
	\section{Conclusions}
	In this paper, we propose an RGCN-based model for semantic parsing of KBQA with extracting global semantics to maximize the performance of question learning. Our approach pays more attention to structure semantics of query graphs extracted by RGCN and enhanced relational semantics in KBQA. In this sense, our approach provides a new way to capture the intention of questions better. In this paper, we generalize our model to support complex questions, such as comparing questions. In future work, we are interested in extending our model for more complex practical questions.

	\clearpage
	
	\newcommand{\etalchar}[1]{$^{#1}$}

\end{document}